\pdfoutput=1
\documentclass{bmcart}

\usepackage{amsthm,amsmath,amsfonts}
\usepackage{comment}
\usepackage{xcolor}
\usepackage{hyperref}
\usepackage{cleveref}
\usepackage{subfigure}
\usepackage[acronym]{glossaries}

\makeglossaries

\newacronym{6d}{6D}{6-Degrees-of-Freedom}
\newacronym{dnn}{DNNs}{Deep Neural Networks}
\newacronym{cnn}{CNNs}{Convolutional Neural Networks}
\newacronym{ep}{EP}{EfficientPose}
\newacronym{lm}{LM}{Linemod}
\newacronym{lm-o}{LM-O}{Linemod-Occluded}
\newacronym{hb}{HB}{HomebrewedDB}
\newacronym{hope}{HOPE}{Houseold Objects for Pose Estimation}
\newacronym{aflm}{AF-LM}{Aruco-Free Linemod}

\usepackage[utf8]{inputenc} 

\usepackage{graphicx}

\startlocaldefs
\endlocaldefs

\begin{document}

\begin{frontmatter}

\begin{fmbox}
\dochead{Research}


\title{Uncovering the Background-Induced bias in RGB based 6-DoF Object Pose Estimation}

\author[
   addressref={aff1, aff2},                   
   email={elena.govi@unimore.it}   
]{\inits{EG}\fnm{Elena} \snm{Govi}}
\author[
   addressref={aff1, aff2},
   email={davide.sapienza@unimore.it}
]{\inits{DS}\fnm{Davide} \snm{Sapienza}}
\author[
   addressref={aff1, aff2},
   email={carmelo.scribano@unimore.it}
]{\inits{CS}\fnm{Carmelo} \snm{Scribano}}
\author[
   addressref={aff1},
   email={257211@studenti.unimore.it}
]{\inits{TP}\fnm{Tobia} \snm{Poppi}}
\author[
   addressref={aff1},
   corref={aff1},
   email={giorgia.franchini@unimore.it}
]{\inits{GF}\fnm{Giorgia} \snm{Franchini}}
\author[
   addressref={aff3},
   email={paola.ardon@tii.ae}
]{\inits{PA}\fnm{Paola} \snm{Ard\'on}}
\author[
   addressref={aff4},
   email={micaela.verucchi@hipert.it}
]{\inits{MV}\fnm{Micaela} \snm{Verucchi}}
\author[
   addressref={aff1, aff4},
   email={ }
]{\inits{MK}\fnm{Marko} \snm{Bertogna}}

\address[id=aff1]{
  \orgname{Department of Physics, Informatics and Mathematics, University of Modena and Reggio Emilia}, 
  \city{Modena},                              
  \cny{IT}                                    
}
\address[id=aff2]{
  \orgname{Department of Mathematical, Physical and Computer Sciences, University of Parma}, 
  \city{Parma},                              
  \cny{IT}                                    
}
\address[id=aff3]{%
  \orgname{TII Technology Innovation Institute},
  \city{Abu Dhabi},
  \cny{UAE}
}
\address[id=aff4]{%
  \orgname{Hipert SRL},
  \city{Modena},
  \cny{IT}
}



\end{fmbox}


\begin{abstractbox}

\begin{abstract} 
 In recent years, there has been a growing trend of using data-driven methods in industrial settings. These kinds of methods often process video images or parts, therefore the integrity of such images is crucial. Sometimes datasets, e.g. consisting of images, can be sophisticated for various reasons. It becomes critical to understand how the manipulation of video and images can impact the effectiveness of a machine learning method. Our case study aims precisely to analyze the Linemod dataset, considered the state of the art in 6D pose estimation context. That dataset presents images accompanied by ArUco markers; it is evident that such markers will not be available in real-world contexts. We analyze how the presence of the markers affects the pose estimation accuracy, and how this bias may be mitigated through data augmentation and other methods.
Our work aims to show how the presence of these markers goes to modify, in the testing phase, the effectiveness of the deep learning method used. In particular, we will demonstrate, through the tool of saliency maps, how the focus of the neural network is captured in part by these ArUco markers. 
Finally, a new dataset, obtained by applying geometric tools to Linemod, will be proposed in order to demonstrate our hypothesis and uncovering the bias. 
Our results demonstrate the potential for bias in 6DOF pose estimation networks, and suggest methods for reducing this bias when training with markers.


\end{abstract}


\begin{keyword}
\kwd{Digital image}
\kwd{Data manipulation}
\kwd{Linemod}
\kwd{Convolutional Neural Networks}
\kwd{Deep Learning explainability}
\kwd{6DoF pose estimation}
\kwd{Saliency maps}
\end{keyword}


\end{abstractbox}
%

\end{frontmatter}




\section{Introduction}
\subsection{Domain of interest}
Recovering the \acrlong{6d} (often referred to as \acrshort{6d}) pose of an object from a single RGB image is a relevant computer vision problem with several applications in the domains of industrial automation \cite{wuest2007tracking}, robotics \cite{zhu2014single} \cite{sapienza2023modelbased}, automotive \cite{chen2017multi} \cite{geiger2012we}, augmented reality \cite{marchand2015pose} and several others.
Generally speaking, given an RGB image $I\in \mathbb{R}^{w\times h\times 3}$, a \acrshort{6d} pose estimation algorithm should recover the translation $\mathbf{t}=(t_x,t_y,t_z)$ and rotation $\mathbf{R}=(r_x,r_y,r_z)$ vectors that describe the position and orientation of an object in the camera coordinate system \cite{shapiro2001computer}. Recent learning-based techniques that use \acrfull{dnn} have proven to achieve very high performance scores in the \acrshort{6d} object pose estimation, achieving state-of-the art results. 
\acrfull{cnn} in particular, achieve impressive results in a wide variety of computer vision problems, including \acrshort{6d} pose estimation from RGB input, at the cost of being extremely data driven and requiring huge amounts of labeled training examples. In \Cref{sec2:1} we present a taxonomy of the most popular Deep Learning methods for \acrshort{6d} object pose estimation. While for other computer vision tasks the dataset acquisition and labeling are easy to obtain, resorting to manual labeling, this is not the case for \acrshort{6d} object pose, since identifying ground truth translations and rotations from real images is not easily feasible for a human annotator. 
The research community has resorted to either relying on large datasets obtained from a photorealistic simulation \cite{hodan2017t} or to smaller datasets of real-world images labeled by relying on fiduciary markers \cite{hinterstoisser2012model, brachmann2014learning}.

\subsection{Annotation induced Bias}


We refer the reader to \Cref{sec2:1} for an overview of the datasets present in the literature. We are particularly interested in real-world datasets that rely on markers to extract the object pose ground truth. We are curious to analyze if the presence of easily recognizable shapes of markers might bring about a bias in the learning procedure, resulting in an improved success in the \acrshort{6d} pose estimation. 

\noindent Commonly used \acrshort{6d} pose estimation methodologies use the \acrfull{lm} dataset \cite{hinterstoisser2012model}. 
This dataset is labelled by fixing the target objects to a rectangular board that is surrounded by ArUco markers \cite{garrido2014automatic}. The ground truth object pose is then retrieved by deploying geometric algorithms which use markers to recover first the board's pose in camera coordinates, and, successively, the object's pose in the same coordinate system.
Utilizing simple image processing algorithms and geometry, the tags are utilized to recover the board's pose in camera coordinates, and finally the object pose is then calculated by applying the known fixed offset from the board's coordinates system to the object's origin.\\

\begin{figure}[!ht]
\centering
    \subfigure{\includegraphics[width=.3\textwidth]{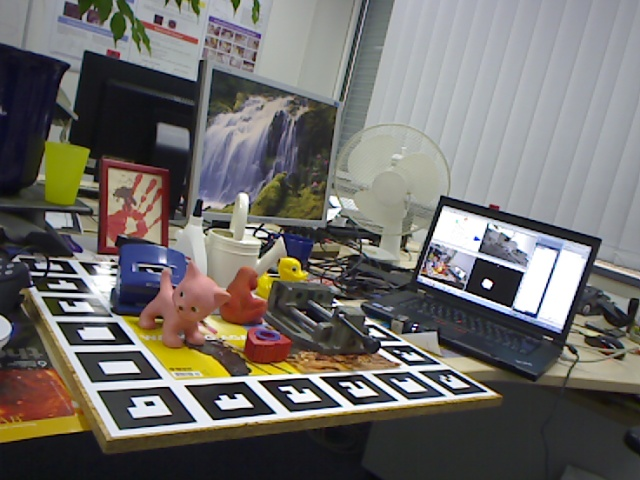}}
     \subfigure{\includegraphics[width=.3\textwidth] {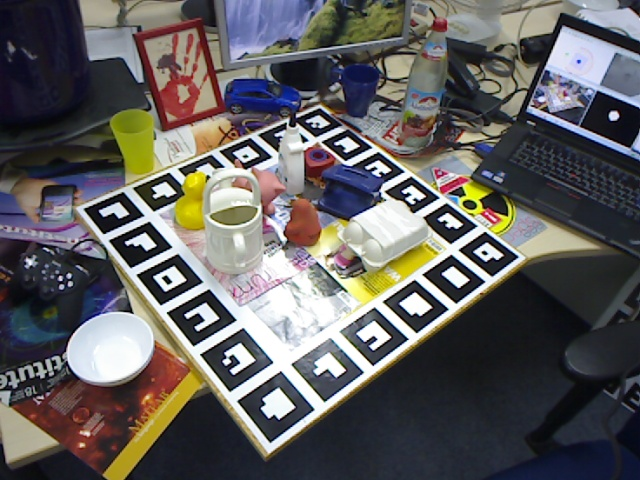}}
     \subfigure{\includegraphics[width=.3\textwidth] {0992.png}}
    \caption{Samples from the Linemod dataset}
    \label{fig:linemod}
\end{figure}

\noindent 

The markers and the board used to recover the ground truth pose are visible in the training and evaluation images. Therefore, we would like to research the effect that the background and markers have on a similar model to the one detailed in \Cref{sec2:3} when predicting the \acrshort{6d} pose from the whole image. Additionally, the arrangement of the objects on the board could cause the model to learn a shortcut or induce unintended behavior. As Linemod is a dataset for single-object, only the pose for the object in the middle of the board is provided for training. We hypothesize that some models could leverage the aspect of other visible objects to infer the \acrshort{6d} pose for the target object in an unintended way.\\
When assessing the efficacy of a proposed method based on Linemod, or similar datasets, it is important to consider its generalization capabilities. This entails evaluating the performance of the method when applied to different scenarios and practical applications, such as robotic manipulation or object tracking for trajectory planning purposes. Furthermore, factors such as changing backgrounds or the lack of ArUco markers or other types of markers should also be taken into account.\\

\noindent To this end, the following factors should be taken into account:

\begin{itemize}
    \item How beneficial is it to have the target object positioned in the center of the board?
    \item To what extent is the method affected by a static or semi-static background? 
    \item Does the network utilize \acrshort{6d} pose information from visible markers and the objects surrounding it?
\end{itemize}

\subsection{Methodology overview}
This paper presents a qualitative and quantitative analysis of \acrfull{ep} \cite{bukschat2020efficientpose}, which was chosen as an ideal candidate for attempting to answer the aforementioned questions due to its state-of-the-art performance on Linemod \footnote{https://paperswithcode.com/sota/6d-pose-estimation-on-linemod} and its ability to operate on the full image. The purpose of this analysis is to illustrate a possible process to assess the generalization properties of a model, which is an essential requirement for any real-world application. 
In addition, this paper aims to emphasize the significance of selecting a proper dataset when training new models. In fact, relying on a dataset which introduces some bias could lead to deceptive outcomes.


\noindent Our work could further be situated within the field of \textit{\acrshort{6d} pose explainability}, an area which, to the best of our knowledge, has not been discussed previously. Given the prevalence of \acrshort{cnn} in computer vision and their tremendous power, it is essential to use \textit{interpretable} models which can explain their predictions. This is important for identifying failure modes, enabling researchers to concentrate on the most promising directions. Furthermore, to ensure the reliability of \acrshort{cnn} in real-world applications, it is necessary to set up appropriate confidence and trust.\\
\noindent\textbf{Paper organization:} In \Cref{sec2},  the state-of-the-art for \acrshort{6d} pose estimation datasets are described, insights into the \acrshort{6d} pose Deep Learning methods, with a focus on EfficientPose, are provided. Additionally, the Saliency maps methods for interpreting the learning process are explained. In \Cref{sec3} are described in details the experiments we have done to demonstrate our hypothesis. In \Cref{sec:4} numerical and visual results are discussed, with their limitations and consequences. Finally, in \Cref{sec:5} there are conclusive considerations.

\section{Related works}\label{sec2}
\subsection{Datasets}\label{sec2:1}
\subsubsection{Linemod and Linemod-Occluded}
\acrlong{lm} \cite{hinterstoisser2012model} is one of the most common benchmarks in the \acrshort{6d} pose estimation domains: it consists of real images with 15 classes (or object models), acquired from different views. A subset of images is provided for each object class, which provides ground truth \acrshort{6d} pose label only for the target object, which is placed around the center of a custom-made work plane and surrounded by other cluttering objects that cause only mild occlusion. As previously noted, the working plane consists of a chessboard-like structure delineated by custom-made ArUco markers. The work of \acrfull{lm-o} \cite{brachmann2014learning} introduces additional ground truth annotations for all modeled objects in one of the test sets, incorporating various levels of occlusion, resulting in a more challenging pose estimation task.

\begin{figure}[h!]
\centering
    \subfigure[T-Less]{\includegraphics[width=.3\textwidth]{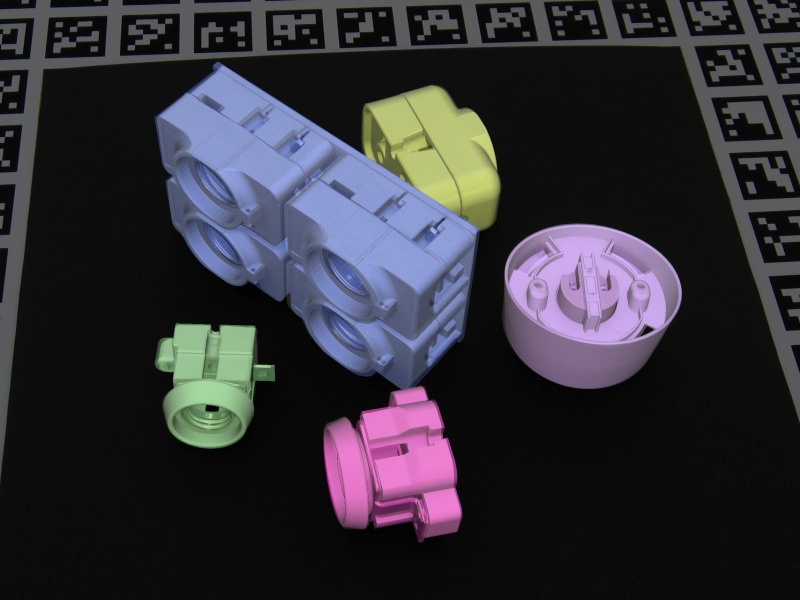}}
     \subfigure[HB]{\includegraphics[width=.3\textwidth] {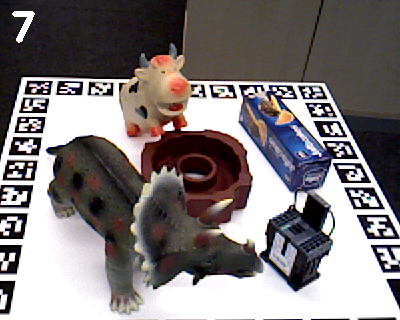}}
     \subfigure[HOPE]{\includegraphics[width=.3\textwidth] {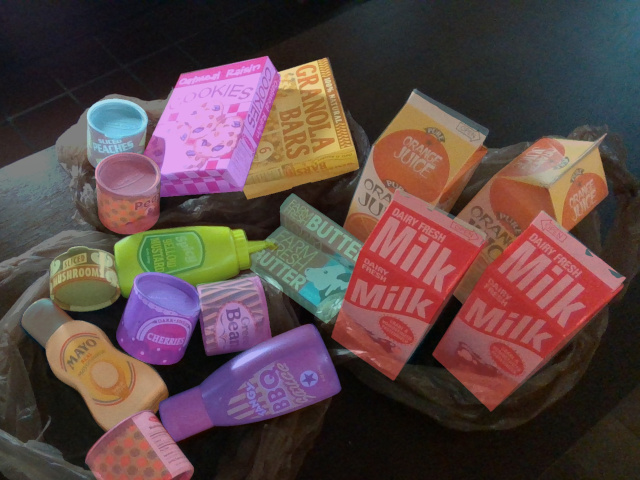}}
    \caption{Samples from various datasets}
    \label{fig:datasets}
\end{figure}

\subsubsection{T-LESS}
The T-LESS dataset was presented by \cite{hodan2017t}, and consists of 30 industry-relevant objects which lack texture or discernible color, as well as 20 RGB-D scenes that were recorded through three synchronized cameras: the PrimeSense Carmine 1.09, the Kinect 2 RGB-D cameras, and the Canon RGB camera. The objects featured in the dataset present some symmetries and mutual similarities, with some being a combination of multiple objects.

\subsubsection{HomebrewedDB}
Structurally similar to T-LESS, \acrfull{hb} \cite{kaskman2019homebreweddb} covers a wider range of objects and provides more challenging occlusions. It consists of 33 highly accurate 3D models of toys, household objects, and low-textured industrial objects of varying sizes, along with 13 sequences containing 1340 frames filmed with two RGB-D sensors. The scenes range from simple (three objects on a plain background) to complex (highly occluded with eight objects and extensive clutter). Interestingly, a chessboard-like pattern similar to the one used in Linemod is clearly visible also in HB.


\subsubsection{HOPE}
NVIDIA \acrfull{hope} \cite{tyree20226} introduced a new dataset of toy grocery objects. The annotations for this dataset were obtained manually, through the identification of point correspondences between images and 3D textured object models. During the acquisition phase, ten different environments, with five object arrangements/camera poses per environment, were used. These 50 different scenes exhibit a wide variety of backgrounds, clutter, poses, and lighting. To provide additional clutter and partial occlusion, objects are also placed in other containers, such as bags or boxes. Of significance in the scope of this paper, the dataset is advantageous as it does not utilize markers or ArUco markers during acquisition. Moreover, the different environments permit to better generalize. In total, the dataset contains 50 unique scenes, 238 images and 914 object poses. Once set the camera and the object position, some light effects are applied, in order to have more images with little differences in shadows and change colors, thereby resulting in more static images that did not need to be annotated.

\subsection{6-DoF Pose Estimation}\label{sec2:2}
Deep-Learning based \acrshort{6d} pose estimation from RGB images can be divided into two approaches: \textit{top-down}, which uses a 2D detector to identify 2D targets (either keypoints or bounding boxes) in the image before estimating the pose of each object, and \textit{bottom-up}, which estimates the \acrshort{6d} pose of all the objects directly. 
The first category includes the keypoint-based methods which, first, extract 2D keypoints from the image, either chosen directly from the object's surface or as 2D projections of the eight cuboid corners, and then solve a Perspective-n-Point (PnP) problem \cite{li2012robust}, \cite{lepetit2009epnp} to recover the \acrshort{6d} pose. 
Top-down methods can also include dense methods \cite{li2019cdpn}, that operate by predicting the corresponding 3D model point from each 2D pixel of the object and then solving the PnP problem from sense 2D-3D correspondences between points.
Bottom-up methods on the other hand in principle could simply regress the \acrshort{6d} pose directly from the image, however, directly estimating the 3D rotation is also difficult, since the non-linearity of the rotation space makes \acrshort{cnn} less generalizable. Recent works \cite{kehl2017ssd, bukschat2020efficientpose} extend single-shoot 2D object detectors to additionally regress translation and rotation vectors for each detected object.

\subsubsection{EfficientPose}\label{sec2:2:1}
EfficientPose \cite{bukschat2020efficientpose} is an extension of a widely used 2D detector, EfficientDet (ED) \cite{tan2020efficientdet}, based on the popular convolutional backbone EfficientNet \cite{tan2019efficientnet}. In a single shot, the architecture is able to predict the class, the 2D bounding box, rotation, and translation of one or more objects, given an RGB image as input.
In detail, two analogous to the classification and bounding box regression subnetworks are added to ED, modeled after the classification and bounding-box regression of the original model. The rotation subnetwork predicts the rotation vector $\mathbf{r}\in \mathbb{R}^3$, in an axis-angle representation. Its architecture is similar to the class and bounding box regression, with the addition of an iterative refinement module.
The final rotation is then the sum $ \mathbf{r} = \mathbf{r}_{init} + \mathbf{\Delta r}$ where $\mathbf{r}_{init}$ is the initial estimate for the rotation, while $\mathbf{\Delta_r}$ is the iterative refinement module, given as output of separable convolutional layers, group normalization and activation functions. The translation network on the other side shares a similar structure. Instead of regressing directly $(t_x, t_y, t_z)$, the t-architecture predicts separately $(c_x, c_y)$, which represents the object center in the image, and $t_z$. After this, $t_x$ and $t_y$ are obtained from $(c_x, c_y)$ and fixed camera parameters, as done in \cite{xiang2017posecnn}.


\subsection{Saliency Maps}\label{sec2:3}
\begin{figure}[h!]
\centering
    \subfigure[Input]{\includegraphics[width=.22\textwidth]{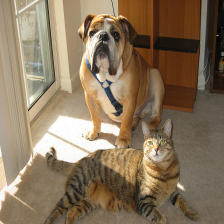}}
     \subfigure[Backprop]{\includegraphics[width=.22\textwidth]{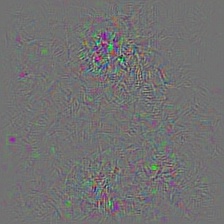}}
     \subfigure[Grad-CAM]{\includegraphics[width=.22\textwidth]{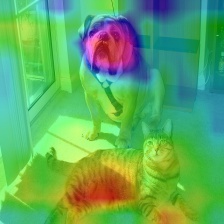}}
     \subfigure[Guided-GC]{\includegraphics[width=.22\textwidth]{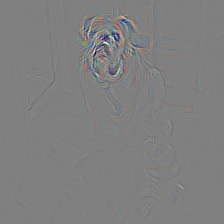}}
    \caption{Output of saliency methods for the class Mastiff. Source \cite{uozbulak_pytorch_vis_2022}}
    \label{fig:saliency_methods}
\end{figure}

Interpretable Machine Learning (IML) \cite{molnar2018guide} has experienced rapid growth in the Machine Learning domain, primarily focusing on elucidating the process behind a model's specific prediction. While some models, such as Decision Trees and Rule Based Classifiers, are inherently interpretable, \acrshort{dnn} cannot be interpreted in the same manner and are generally considered to be a black-box predictor after training. Pixel attribution is a family of attribution methods designed to interpret the prediction of models that operate on images, with the aim to produce a saliency map which highlights the importance of individual (or groups of) pixels in the input image for the model's specific output. One of the simple methods to generate saliency maps, introduced in \cite{simonyan2013deep} is sometimes referred to as Vanilla-Gradient, and it consists of computing the gradient of the output prediction with respect to the input image. More formally, given $s=\Psi(I)$, where $\Psi(\cdot)$ denotes the (trained) neural network, the gradient $\frac{\partial \Psi}{\partial I}$ can be easily obtained via standard backpropagation. Grad-CAM \cite{selvaraju2017grad} is a more recent method to produce saliency maps: differently from the vanilla gradient, the gradient of the output $y_s$, that is the class score before the softmax, is computed with respect to the output feature map $A^k$ of a convolutional layer (ideally, the last one before the global average pooling of traditional classification models). The gradient tensor is then averaged across the spatial dimensions indexed by $i,j$ to obtain a single vector $\alpha^s_k\in R^{c}$ where $c$ denotes the number of channels: 
 

\begin{equation}\label{eq:1}
    \alpha^s_k = \frac{1}{Z}\sum_{i}\sum_{j}\frac{\partial y_s}{\partial A_{ij}^k}
\end{equation}

\noindent Finally, the saliency maps are obtained by weighting each channel of the feature map $A^k$ with the corresponding value of $\alpha_k$:
\begin{equation}\label{eq:2}
    L_{GC} = ReLU\left(\sum_{k}a_kA^k\right)
\end{equation}

We refer the interested reader to the exceptional work of \cite{molnar2018guide} for a deeper understanding of the discussed methods and several others, which falls beyond the scope of this work. In \Cref{3:2} we will discuss our generalization of Grad-CAM for the regression problem of our interest.

\section{Methodology}\label{sec3}
The aim of this research is to explore the potential for bias to be introduced into the model due to the presence of visible artifacts, particularly markers employed in the data collection process. We will focus our experimentation on the Linemod dataset, a well-known non-synthetic dataset, and the EfficientPose model, which is currently the most effective fully-convolutional network for Linemod. We propose a mixed evaluation of qualitative observations, utilizing the attribution methods discussed in \Cref{sec2:3}, and quantitative experimentation, involving the evaluation of results obtained from modified versions of the Linemod dataset, to assess the validity of our hypothesis.\\

\noindent The proposed methodology, can be briefly summarized as follows:
\begin{itemize}
    \item We propose to pre-process Linemod by deliberately masking the visible markers and use the new version of the dataset for the training of EfficientPose. The \acrshort{6d} pose estimation task is then compared between the late and the original training of \acrshort{ep}.
    \item We also introduce a generalization of pixel attribution methods for a regression problem, and show that the saliency maps produced with the extended version of Grad-CAM support our hypothesis that the model's predictions are contingent upon the presence of fiduciary markers on Linemod.
\end{itemize}

\noindent In the remainder of this section we detail the building blocks of the proposed methodology, then in \Cref{sec:4} the conducted analysis is proposed and discussed in detail.

\subsection{Dataset Masking strategy}\label{sec3:1}

For our analysis, two datasets have been used. One is the original Linemod dataset. The other is a modified Linemod Dataset, where ArUco markers are covered with black squares. The process of deleting ArUco markers is based on geometrical tools and on the objects' ground truth. 
\begin{figure}[h!]
\centering
    \subfigure{\includegraphics[width=.3\textwidth]{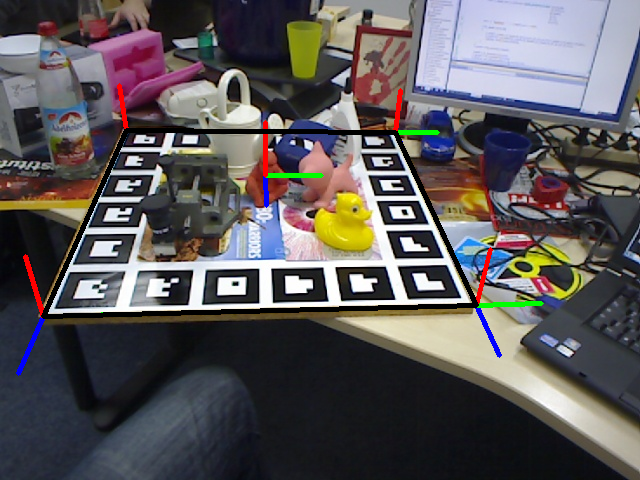}}
     \subfigure{\includegraphics[width=.3\textwidth] {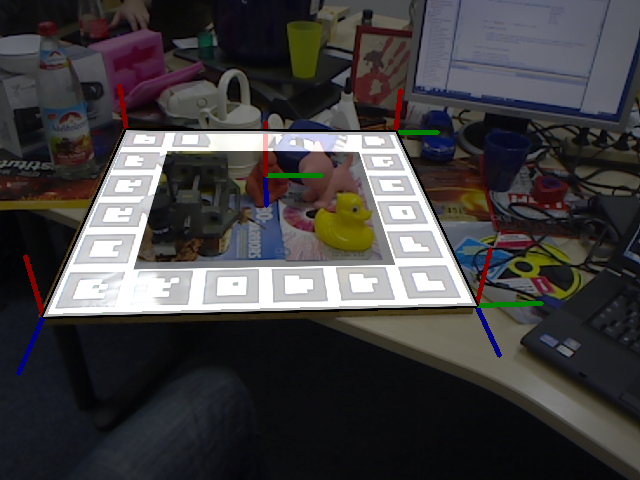}}
     \subfigure{\includegraphics[width=.3\textwidth] {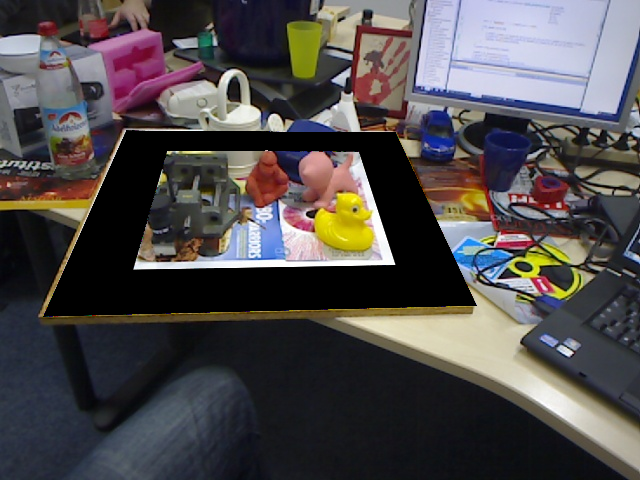}}
    \caption{Three steps of our geometric procedure in order to remove ArUco markers. }
    \label{fig:freeArUco}
\end{figure}
Considering the 3D origin point $[0, 0, 0]$ as the object's pivot, when no translation and rotation are applied, the positions of the black square's four corners can be identified. The four corners correspond to the upper-left, upper-right, lower-left and lower-right corners of the ArUco chessboard. Each corner is represented by a 3D point defined as $[ \pm \delta_x,\pm  \delta_y, \pm  \delta_z ]$, where $\delta_x$, $\delta_y$, $\delta_z$ are the axis offsets between the corners and the origin. 
Corners are identified only once for each object. Then, for each object image, the corresponding rotation and translation are applied to accurately project the black square at its correct image position (see Figure \ref{fig:freeArUco}). \acrshort{aflm} Dataset has been done in order to show and discuss how they influence final results.

\begin{figure}[h!]
    \centering
    \begin{tabular}{ccc}
      \includegraphics[width=.3 \textwidth]{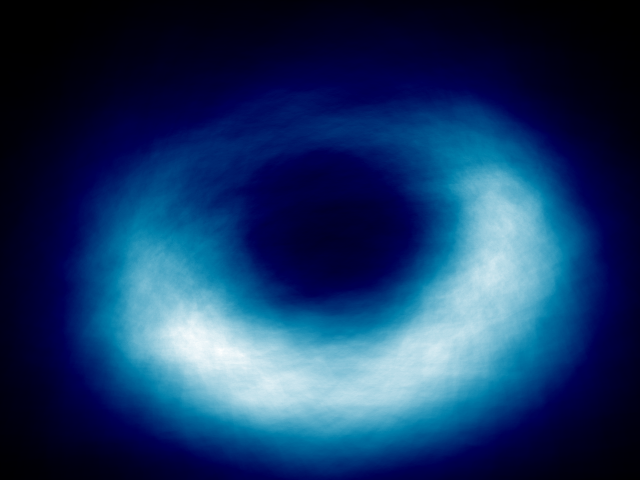}   &   \includegraphics[width=.3 \textwidth]{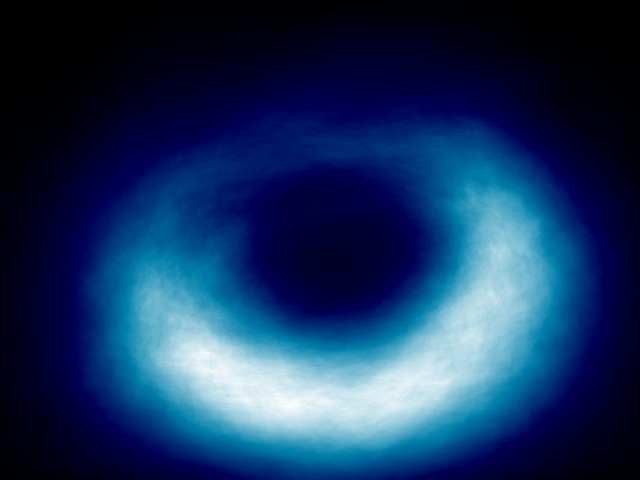} &  \includegraphics[width=.3 \textwidth]{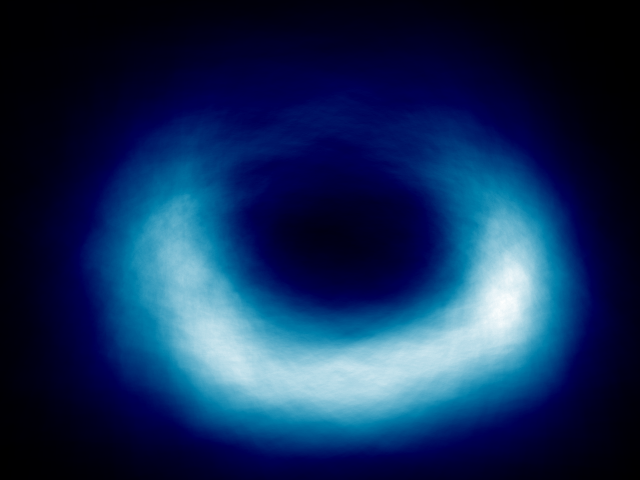} 
    \end{tabular}
    \caption{Density maps based on masks. From left to right: object 1, object 5, object 11. }
    \label{fig:density_maps}
\end{figure}

\noindent In addition, ArUco masks were used also for computing a density map, able to highlight their presence. The images in Figure \ref{fig:density_maps} indicate for each object which areas are ArUco markers concentrations. It is evident that these markers are not equally distributed, instead they focus on the same areas.

\subsection{Saliency maps}\label{3:2}
For the qualitative analysis, inspired by the domain of interpretable AI, we sought to interpret and extract information about how the network learns. This is a fundamental phase in order to assess the capabilities of \acrshort{6d} methods in other real-world scenarios, since metrics are limited in providing insight into a network's comprehension.\\
Saliency methods are mainly developed for multi-class classification problems \cite{russakovsky2015imagenet}, and are thus sometimes referred to as Class Activation Maps (CAM) methods. At inference time, classification models output a probability score for each possible class, and the highest scoring class is selected as the predicted one. However, in this study, we focus on a regression problem: this is pertinent as the gradient-based methods that we plan to use, Vanilla Saliency \cite{simonyan2013deep} and Grad-CAM \cite{selvaraju2017grad}, suppress the negative part of the gradient since it corresponds to a decrease in the score for the class of interest. In a scenario similar to ours (i.e, regression of rotation values in $[-\pi,\pi]$) we are clearly interested only in the magnitude of the gradient. Hereafter, we formalize the exact formulation of the saliency methods used in our work.

\paragraph{Vanilla Saliency} It is straightforward to adapt the vanilla saliency for the regression. The Rotation head of EfficientPose outputs a tensor $\overline{R}\in\mathbb{R}^{N \times 3}$ of $N$ candidate regressions. In a single-object scenario, as in our case, the rotation vector $\mathbf{r}$ for the target object is recovered as the one with the highest confidence. An identical approach can be adopted also for the translation regression. The gradient of $\mathbf{r}$ computed with respect to the input image $I$ is a tensor with the same shape as the input image, which is reduced to a single channel by averaging. Unlike the original formulation, we retain both the negative and positive values of the gradient for the reasons mentioned above. For visualization, the saliency map (single-channel) is normalized to the interval $[0,255]$ using min-max.

\begin{figure}[!ht]
\centering
    \includegraphics[width=0.95\textwidth]{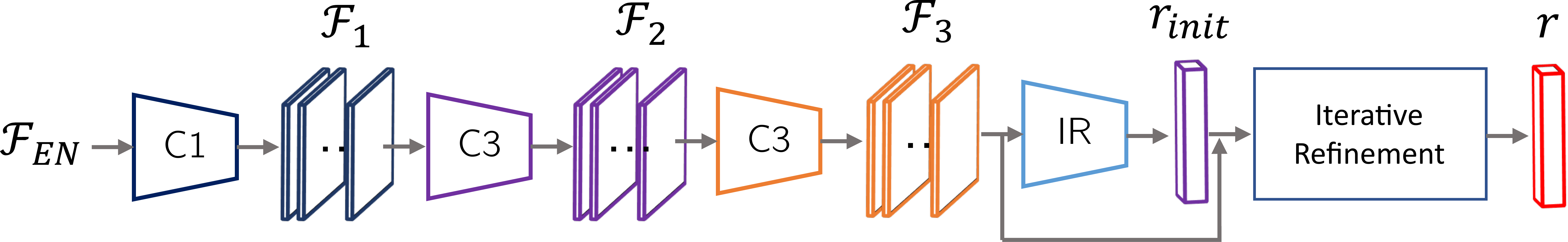}
    \caption{Rotation subnetwork for EfficientPose. The Translation subnetworks share an identical structure.}
    \label{fig:ep_rot}
\end{figure}

\paragraph{Grad-CAM} Adapting Grad-CAM to our problem is far more challenging. The original formulation operates on the feature map produced by the last convolutional layer of a classification model; however, the structure of the regression head of \acrshort{ep} (simplified in \Cref{fig:ep_rot}) makes it difficult to choose the correct feature map. Therefore, we opted to use a feature map obtained as the combination of the three convolutional layers that precede the output of the initial prediction $\overline{r}_{init}$, as the refinement module is used to predict small additive offsets to the initial prediction, which may be Identity mappings if the initial prediction does not require refinement.

Let $\mathcal{F}_1$, $\mathcal{F}_2$ and $\mathcal{F}_3 \in \mathbb{R}^{\overline{W}\times\overline{H}\times\overline{C}}$, with $\overline{W},\overline{H},\overline{C} \in \mathbb{N}$ be the three intermediate feature maps, we construct an aggregated featuremap $\mathcal{F}_t \in \mathbb{R}^{\overline{W}\times\overline{H}\times3\overline{C}} = \{\mathcal{F}_1\oplus\mathcal{F}_2\oplus\mathcal{F}_3\}$ where $\oplus$ denotes the concatenation on the channel axis. \Cref{eq:1} is adapted by replacing $A^k$ with $\mathcal{F}_t$ and replacing the average pooling with $L_2$ norm, in order to avoid opposing gradient values to elide each other in the sum. The new formulation for the pooled gradient becomes:
\begin{equation*}
    \alpha_t = \sqrt{\sum_{i=1}^{\overline{W}}\sum_{j=1}^{\overline{H}}\left(\frac{\partial\mathbf{r}}{\partial \mathcal{F}_t^{ij}}\right)^2} \in \mathbb{R}^{3\overline{C}}
\end{equation*}

\noindent The computation of the saliency maps is obtained by weighting each channel of $\mathcal{F_t}$ with the corresponding value of $\alpha_k$ and accumulating over the channel axis to obtain a single matrix. Differently from \cref{eq:2} the ReLU is removed and the absolute value of the weighed feature map is instead taken:

\begin{equation*}
    L_{GC} = \sum_{t=1}^{3\overline{C}}|a_t\mathcal{F}_t|
\end{equation*}

\noindent The produced saliency map is normalized and interpolated to the input image shape for visualization. The formulation, derived from simple but solid mathematical observations, results in an ideal generalization of Grad-CAM to our case.

\subsection{Evaluated Task metrics}\label{sec3:3}
The metrics that we used to test the networks are two:
\begin{itemize}
\item \textbf{ADD}: the average distance computes the mean distance between each point of the 3D model obtained by the pose matrices $\hat{\mathbf{P}}=[\mathbf{R_{est},t_{est}; 0},1]$ and $\overline{\mathbf{P}}=[\mathbf{R_{gt},t_{gt}; 0},1]$

Given the model $\mathcal{M}$, the estimated pose $\hat{\mathbf{P}}$ and the ground truth $\mathbf{\overline{P}}$ \[ e_{ADD} = avg_{x\in \mathcal{M}} \Vert \mathbf{\hat{P}x}-\mathbf{\overline{P}x} \Vert\]
      
\item \textbf{ADD-S}: the average closest point distance computes the mean distance between each point of the 3D estimated model and its closest neighbor on the ground truth model: \[ e_{ADD-S} = avg_{\mathbf{x}_{1}\in \mathcal{M}} \min_{\mathbf{x_2} \in \mathcal{M}} \Vert \mathbf{\hat{P}x_1}-\mathbf{\overline{P}x_2} \Vert\]
It is preferred if the model $\mathcal{M}$ has indistinguishable views.
\item \textbf{Criterion of Correctness} The estimated pose is considered correct if $e < \theta_{AD} = k_m d$ where $k_m$ constant generally equal to $0.1$, $d$ = object diameter
\end{itemize}

\subsection{Proposed Evaluations}
For our experimental analysis we focus on the Rotation subtask of \acrshort{ep}, since the translation regression is based on an identical structure, as introduced in \cref{sec2:2:1}, the same principles are agnostic to the regression subtask. We compare the results, both qualitatively and quantitatively, obtained by two architecturally identical instances of \acrshort{ep}: the original model trained on the official Linemod dataset and an alternative version trained on the ArUco-Free dataset introduced in \Cref{sec3:1}. Since we focus on single-class \acrshort{ep}, the evaluation procedure is performed independently for three distinct representative object classes from Linemod.\\

\noindent To summarize, a total of 6 distinct versions of \acrshort{ep} are trained: we pick the standard version of \acrshort{ep} ($\phi=0$) and train on subsets of objects 1 (Ape), 5 (Can) and 11 (Glue) of both original (\acrshort{lm}) and \acrfull{aflm} datasets. For each trained model, we compute \acrshort{6d}-pose metrics (\cref{sec3:3}) and saliency maps (\cref{3:2}) on both the validation subsets of \acrshort{lm} and \acrshort{aflm} for the corresponding object. In the following section we provide the complete analysis of the proposed study.

\section{Results and Discussion}\label{sec:4}
\subsection{Quantitative analysis}

We considered three objects: one with symmetric views (object 11, glue) and the other two asymmetric (object 1, ape and object 5, can). Considering ADD(-S) as a metric, the following rule is used:
\[
ADD(-S)=\begin{cases} ADD-S, & \mbox{if }obj\mbox{ sym }\\ADD, & \mbox{if }obj\mbox{ asym}
\end{cases}\]

\subsubsection{Object 1: the ape}
\begin{figure}
\begin{tabular}{ccc}
\includegraphics[width=.28\textwidth]{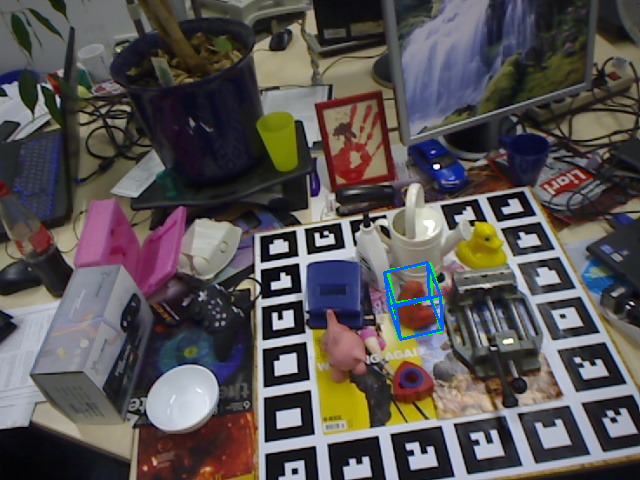} & \includegraphics[width=.28\textwidth]{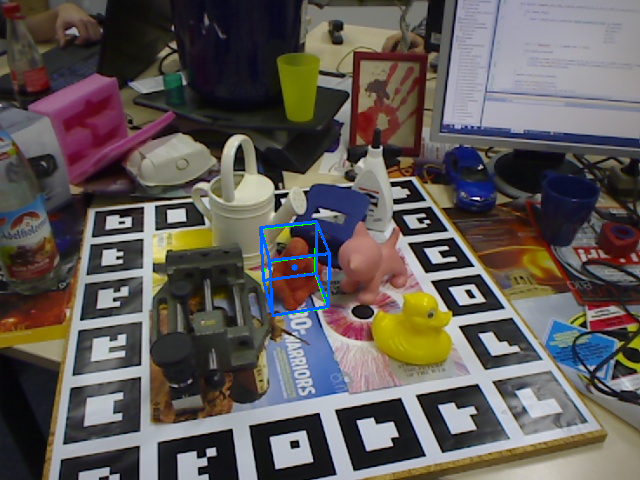} & \includegraphics[width=.28\textwidth]{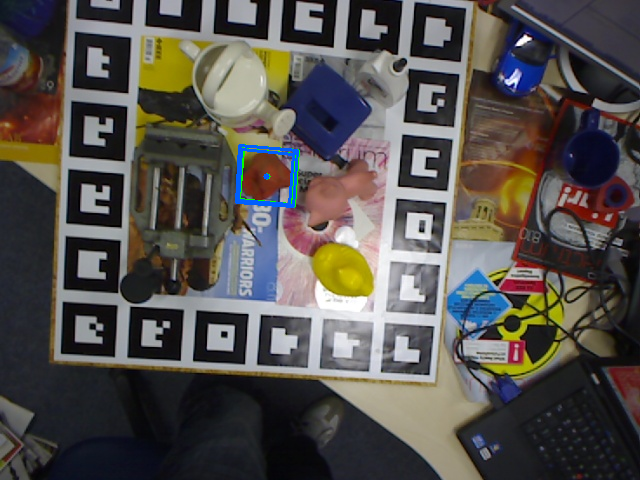} \\

      \includegraphics[width=.28\textwidth]{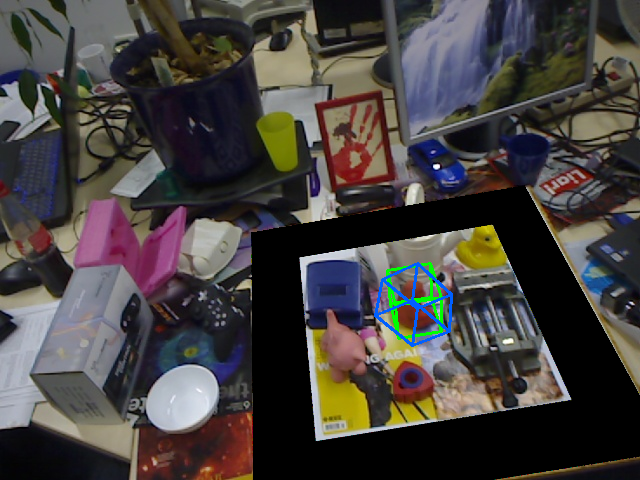} & \includegraphics[width=.28\textwidth]{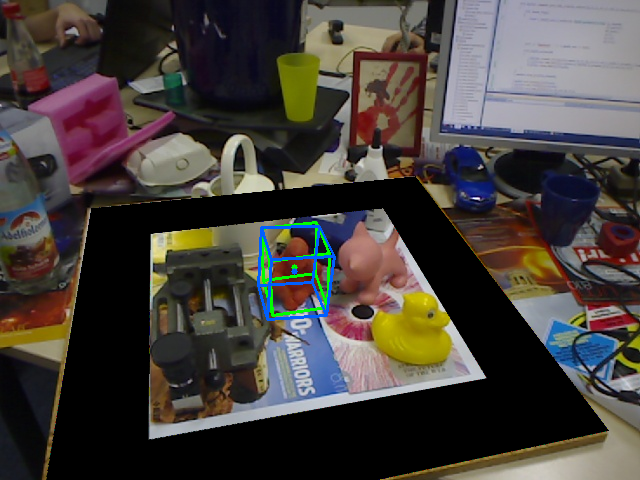} & \includegraphics[width=.28\textwidth]{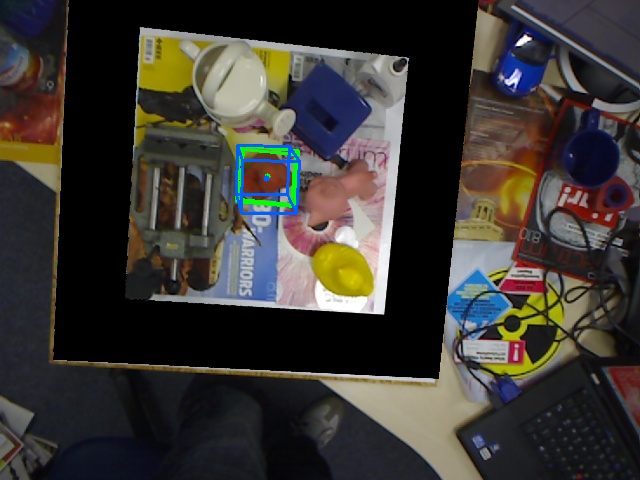} \\

\includegraphics[width=.28\textwidth]{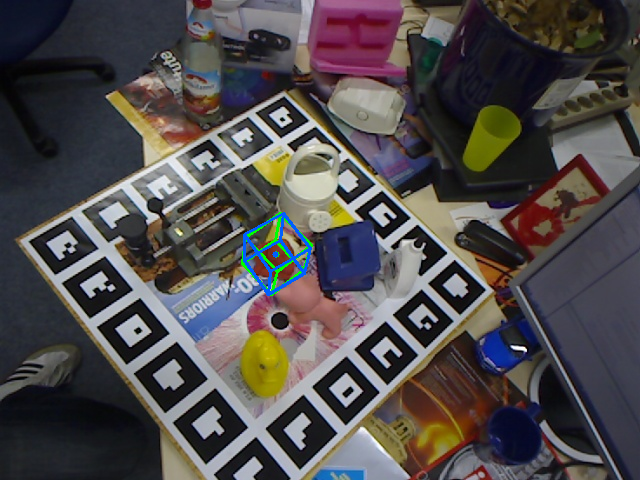}
& \includegraphics[width=.28\textwidth]{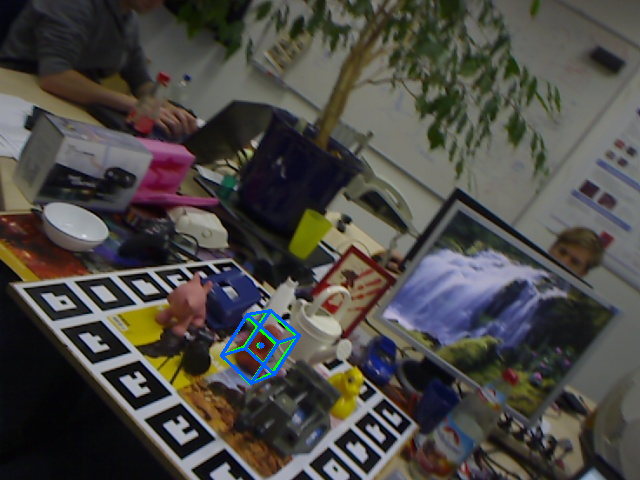}
& \includegraphics[width=.28\textwidth]{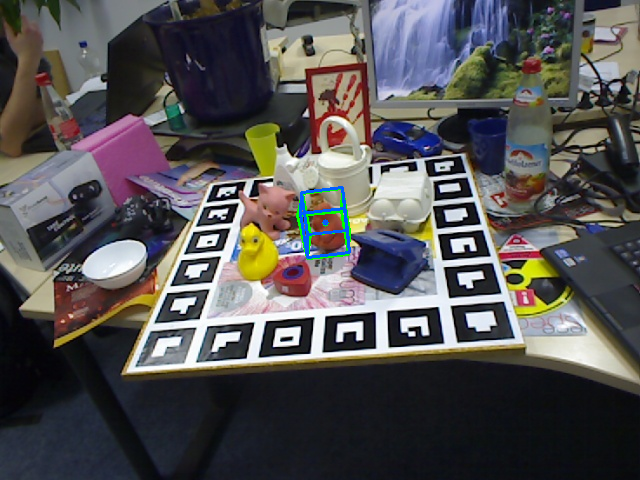}\\

\includegraphics[width=.28\textwidth]{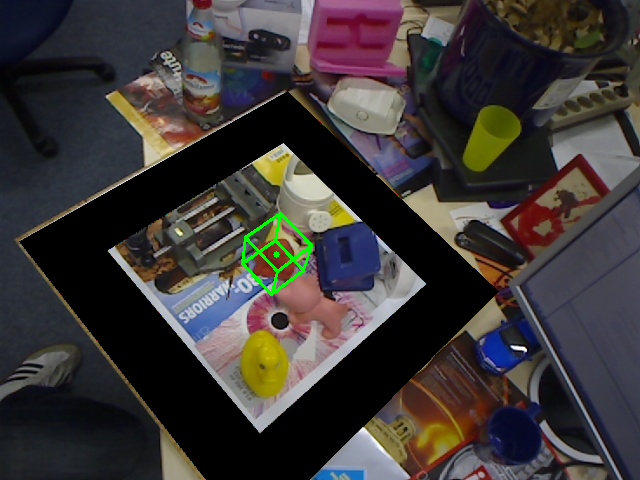} 
& \includegraphics[width=.28\textwidth]{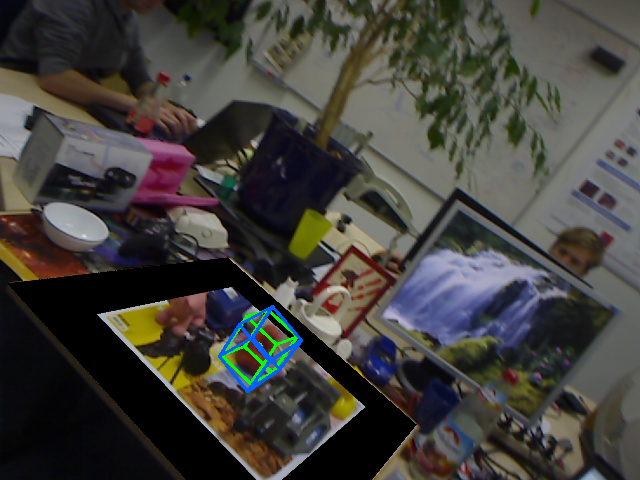}
      & \includegraphics[width=.28\textwidth]{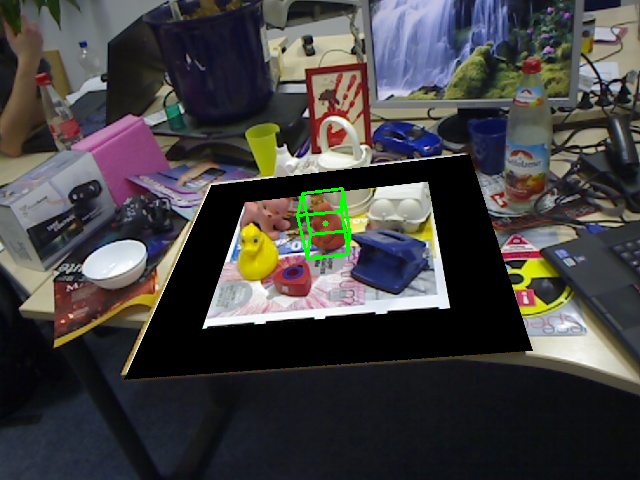}

\end{tabular}
    \caption{These images are used during the test phase on object 1 with weights downloaded from the available Efficient Pose training ($\phi=0$). The green box represents the ground truth pose, while the blue one represents the estimated pose. On the first and third rows there are test images from the original Linemod dataset, while on the second and fourth rows there are the correspondent images from ArUco-Free dataset. We observed that the boxes are all wrong and, in some cases, the network doesn't even detect the object. }
    \label{obj_1_yesArUco}
\end{figure}

Table \ref{obj1_metrics} shows how EfficientPose performs on the two different datasets. We can observe that in both training (with and without ArUco markers) the network learns from the background. In fact, if the test dataset is different from the train one, the pose estimation accuracy collapses. Figure \ref{obj_1_yesArUco} compares estimated (blue) with ground truth (green) bounding boxes, with weights learned on the Original \acrshort{lm} Dataset. When ArUco-Free Dataset is used as test, in some cases the rotation is wrong, in others the object is not even detected. Probably, when the ArUco markers are covered, the network still learns from the black square around the object. 
Object 1 achieves the worst performances on both datasets. 

\begin{table}[h]
    \begin{tabular}{ |c||c|c|c|  }
 \hline
  \multicolumn{4}{|c|}{Test} \\
    \hline
     \scriptsize{Original \acrshort{lm} training}   & \begin{tabular}{c}
          \scriptsize{Original \acrshort{lm}}  \\
            $0.8771$   \end{tabular} & \begin{tabular}{c}
          \scriptsize{\acrshort{aflm}} \\ $\mathbf{0.0162}$   \end{tabular} & $\mathbf{0.4467}$ \\
     \hline
    \scriptsize{\acrshort{aflm} training} & \begin{tabular}{c}
          \scriptsize{\acrshort{aflm}} \\  $\mathbf{0.8848}$   \end{tabular}  & \begin{tabular}{c}
          \scriptsize{Original \acrshort{lm}}  \\
            $0.0$     \end{tabular} & $0.4424$ \\
         \hline
    \end{tabular}
    \caption{This is object 1. ADD is computed on the two datasets with the two types of weights available (trained with $\phi=0$ ). Then, in the right column, for each training, the two test results' averages are computed, in order to observe which experiment performs better both on the test images of the same image type seen during training, and on the test images of the type never seen. }
    \label{obj1_metrics}
\end{table}

\subsubsection{Object 5: the can}
For object 5, the can, we used trained EfficientPose weights with $\phi=0$ on the Original \acrshort{lm} Dataset and on the ArUco-Free Dataset. The object is not symmetric, therefore ADD is used.
For this object, we noticed that the results are similar to the ones of object 1, and confirm our thesis of background-induced bias. However, in this case, EfficientPose with our ArUco-Free Dataset performs better in both cases (Original \acrshort{lm} and \acrshort{aflm} datasets), as shown in Table \ref{obj5_metrics}.
\begin{table}[h]
    \centering
    \begin{tabular}{|c||c|c|c|  }
 \hline
 \multicolumn{4}{|c|}{Test} \\
    \hline

     \scriptsize{Original \acrshort{lm} training}    &  \begin{tabular}{c}
          \scriptsize{Original \acrshort{lm}}  \\
            $0.9852$ \end{tabular} & \begin{tabular}{c}
          \scriptsize{\acrshort{aflm}} \\  $0.0315$ \end{tabular} & $0.5083$ \\
     \hline

    \scriptsize{\acrshort{aflm} training}   &  \begin{tabular}{c}
          \scriptsize{\acrshort{aflm}} \\  $\mathbf{0.9921}$  \end{tabular}   & \begin{tabular}{c}
          \scriptsize{Original \acrshort{lm}}  \\
           $\mathbf{0.1673}$ \end{tabular} & $\mathbf{0.5797}$ \\
         \hline

    \end{tabular}
    \caption{This is object 5. ADD is computed on the two datasets with the two types of weights available(trained with $\phi=0$ ). 
    Then, in the right column, for each training, the averages of the two test results are computed, in order to observe which experiment performs better both on the test images of the same type of the images seen during training, and on the test images of the type never seen.}
    \label{obj5_metrics}
\end{table}

\subsubsection{Object 11: the glue}
The third object we chose for our study is a symmetric object, for this reason, we show ADD-S results in Table \ref{obj11_metrics}. They are higher since ADD-S has more relaxed constraints than ADD (as can be deduced by the definition). In this case, while performances on the same dataset of the training are almost perfect, with an accuracy of $100\%$, the training without ArUco performs better than the other with the opposite dataset. Therefore, the average accuracy is better. 
From these metrics it is evident that the network learns from the background. However, accuracies don't give us information about which background areas are more relevant than others. To explain better these results, we used saliency maps in the next section.

 

\begin{table}[h]
    \centering
\begin{tabular}{|c||c|c|c|}
 \hline
 \multicolumn{4}{|c|}{Test} \\
  \hline
  
     \scriptsize{Original \acrshort{lm} training} &  \begin{tabular}{c}
          \scriptsize{Original \acrshort{lm}}  \\
          
           $\mathbf{1.0000}$
     \end{tabular} & \begin{tabular}{c}
          \scriptsize{\acrshort{aflm}}  \\
          $0.3031$ \end{tabular}  & $0.6516$ \\
     \hline
    \scriptsize{\acrshort{aflm} training}  & \begin{tabular}{c}
          \scriptsize{\acrshort{aflm}}  \\
           $\mathbf{1.0000}$   \end{tabular}  & \begin{tabular}{c}
          \scriptsize{Original \acrshort{lm}}  \\
           $\mathbf{0.4083}$ \end{tabular} & $\mathbf{0.7042}$ \\
         \hline
    \end{tabular}
    \caption{This is object 11. ADD-S is computed on the two datasets with the two types of weights available (trained with $\phi=0$ ). Then, in the right column, for each training, the two test results' averages are computed, in order to observe which experiment performs better both on the test images of the same image type seen during training, and on the test images of the type never seen.}
    \label{obj11_metrics}
\end{table}

\subsection{Qualitative analysis}

\begin{figure}
    \centering
    \begin{tabular}{c}
             \includegraphics[width=.9 \textwidth]{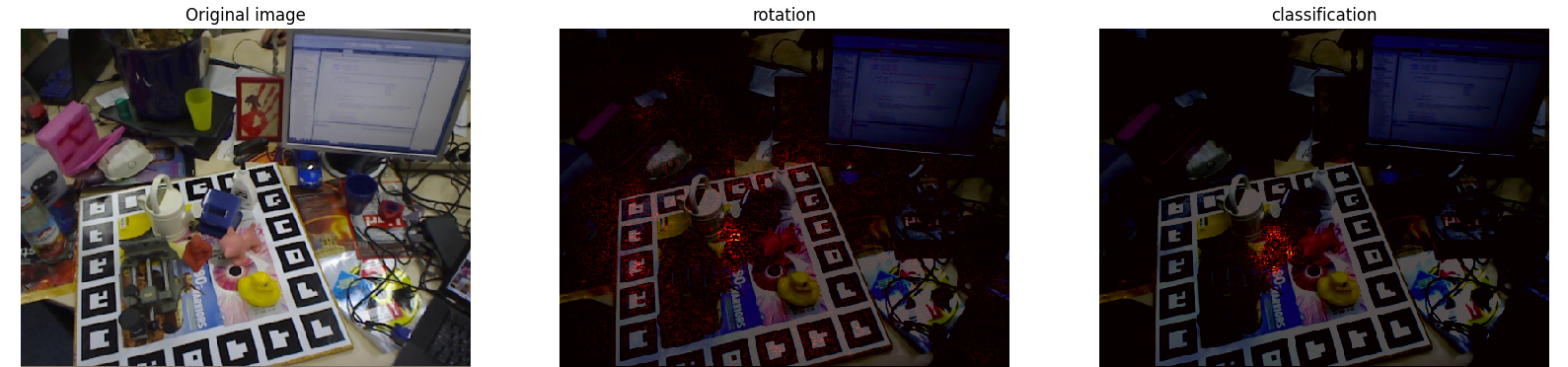}
    \end{tabular}

    \caption{Vanilla Saliency applied on two images of the Original LineMod dataset. Weights are trained on the same dataset. The gradient is computed given two different outputs: in the second image it is based on the regressor, in the third image it is based on the classificator. }
    \label{original_saliency}
\end{figure}

\begin{figure}
\begin{tabular}{lll}
\includegraphics[width=0.28 \textwidth]{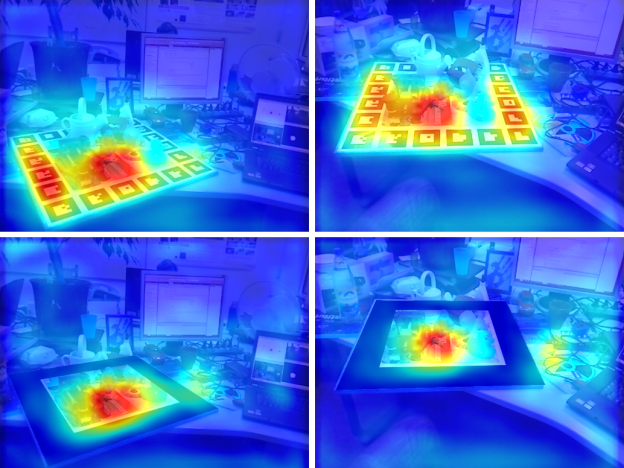} & \includegraphics[width=0.28 \textwidth]{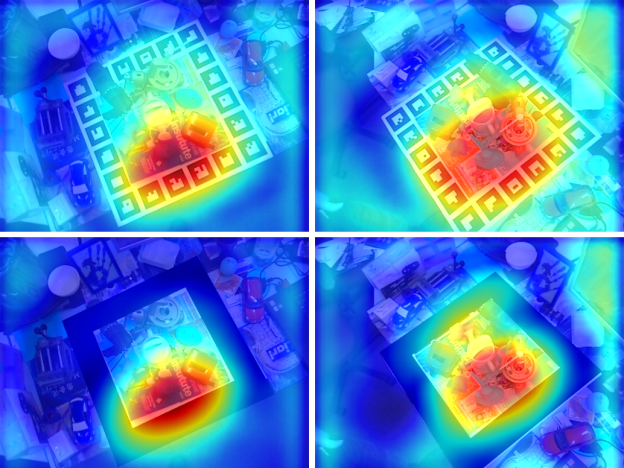} & \includegraphics[width=0.28 \textwidth]{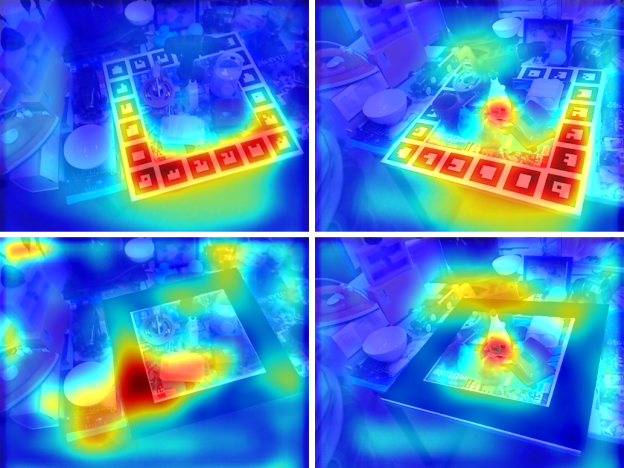} \\
\end{tabular}
    \caption{On the top: saliency maps with weights provided by EfficientPose on the original Linemod Dataset.  On the bottom: saliency maps with the same weights, applied to our ArUco-Free Linemod dataset.
    From left to right: object 1, object 5, object 11.}
    \label{saliency_trainArUco}
\end{figure}

\begin{figure}
\begin{tabular}{lll}
\includegraphics[width=0.28 \textwidth]{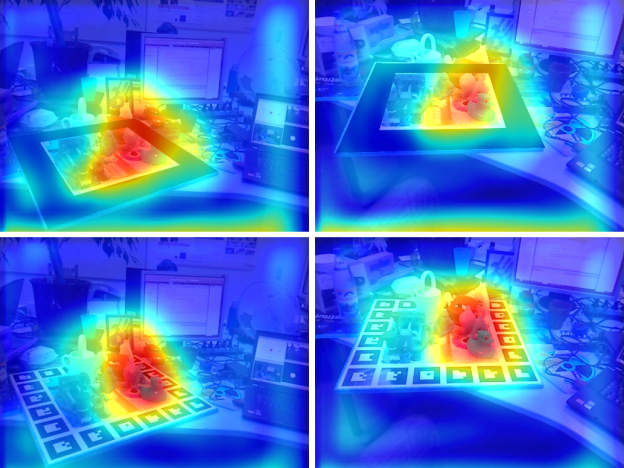} & \includegraphics[width=0.28 \textwidth]{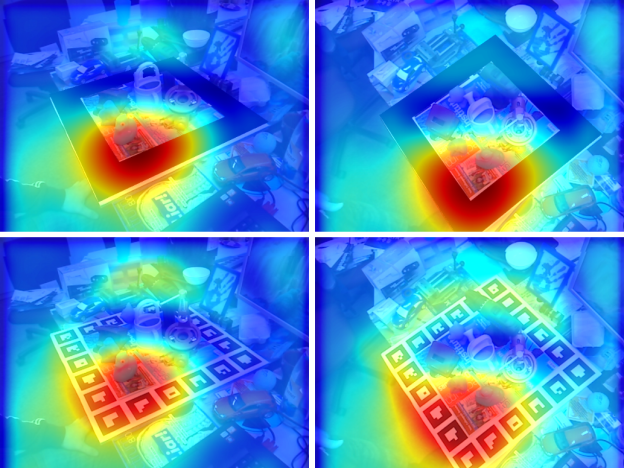} & \includegraphics[width=0.28 \textwidth]{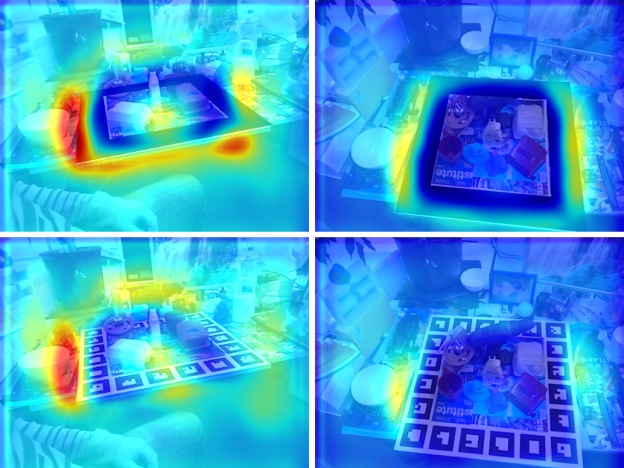} \\
\end{tabular}
    \caption{On the bottom: saliency maps with weights learned on the ArUco-Free Linemod Dataset, applied to the Original \acrshort{lm} dataset. 
    On the top: saliency maps with the same weights, applied to our ArUco-Free Linemod dataset. 
    From left to right: object 1, object 5, object 11.}
    \label{saliency_train_noaruco}
\end{figure}

Figure \ref{original_saliency} shows Vanilla Saliency maps. They are computed with respect to two Original \acrshort{lm} images. The weights are given by EfficientPose code for object 1, with $\phi=0$. The map represents the gradient magnitude for each pixel. Since \acrshort{ep} has a first common feature extraction phase and, then, is divided into different subnetworks with their own output (classification, bounding boxes, rotation, and translation), two saliencies for each image have been compared. The 'rotation' image represents vanilla saliency map based on the rotation subnetwork, while the 'classification' image comes from the classification subnetwork.
The differences for object 1 in this image are significant. In fact, while the saliency for classification is grouped into the principal object, the ape, instead the rotation saliency is more scattered and goes also on the marker chessboard. It probably means what we expected: that, for the position output, a bias is induced by background, which plays a fundamental role. In order to improve our study with a more sophisticated and explainable saliency method, we also computed saliency maps with GradCAM method. Moreover, sometimes Vanilla Saliency could have a saturation problem, as shown by \cite{selvaraju2017grad} and \cite{molnar2018guide}.

\noindent In Figure \ref{saliency_trainArUco}, we can observe different results for the three objects. The weights are learned on the Original \acrshort{lm}, while the tests are computed on both datasets.
For example, for object 1 (on the left), pixels with higher saliency values are, on the Original \acrshort{lm} (top), distributed on the object and the markers, whereas on the \acrshort{aflm}, they do not put focus on markers, in fact they are covered. It means that, when ArUco values collapse to zero, they are not anymore interesting for the rotation estimation. 

\noindent For object 5, in the center, the focus is on a large area which includes also the can. Therefore, the network uses not only the principal object, but also the background one, to estimate the final pose. This is possible since background objects do not change their position with respect to the can. 

\noindent Similar to the ape behavior, for object 11's pose estimation \acrshort{ep} focuses on marker chessboard, not even observing the principal object. When markers are zeroed, saliency is distributed in not well-defined areas. 
These images prove the fact that there is a background-bias during EfficientPose training with ArUco markers.

\noindent Figure \ref{saliency_train_noaruco} represents saliency maps obtained from our weigths, learned on the \acrshort{aflm}. These results are quite different from the previous ones.

\noindent For object 1 the saliency is not on the markers, however, the network doesn't focus on the ape, looking for information in other background objects. Saliency does not change a lot: it means that markers are not the central keypoints for pose estimation. Nevertheless, the network on the Original \acrshort{lm} dataset has the worst performance accuracy.

\noindent Also, in object 5 (in the center) images saliency maps seem to remain the same for both tests. 

\noindent An interesting phenomenon happens with object 11 (on the right). The most plausible interpretation is that weights on the ArUco-Free dataset predict the pose based on the square edges. When these edges are not so strongly highlighted due to the presence of ArUco markers, the prediction is wrong. For this reason, we think that covering the ArUco markers is useful for uncovering the bias, but it's not enough for obtaining a more generalized dataset.

\noindent An interesting result emerges from the saliency maps calculated for the model trained on the Original \acrshort{lm} dataset. For the original Linemod images, the saliency maps focus on both the ArUco markers and the area close to the target object, as described above and depicted in Figure \ref{saliency_trainArUco} (top). In contrast, for the ArUco-Free Linemod images, the saliency maps are sometimes zeroed out. This observation can be explained by the fact that, for ArUco-Free Linemod images, some pixels were zeroed for each RGB channel. The zeroing of these pixels, which refer to the ArUco markers considered essential for the \acrshort{ep} model, causes the gradients to be zeroed as well.
It is as if these null values made that part of the hyperplane on which the gradients were calculated constant.

Indeed, the gradient is zero for any constant value, as well as for a minimum or maximum point. 
Furthermore, for these specific images, we verified that the last activations before the last convolutional layer used for the Grad-CAM calculation were not null. 


 \subsection{Consequences}
Finally, this paper aims to give some advices for avoiding background-induced bias in \acrshort{6d} object pose estimation. 
Foremost, networks are preferable which first detect the object, and then predict the pose of the cropped image. Secondly, the dataset choice is also fundamental. 
To avoid biasing the learning method, it is important to change the background from training frames.
In particular, different lights, views and background objects are helpful to overcome this issue. For example, HOPE \cite{tyree20226}, as written in section \ref{sec2:1}, presents images in different scenarios, changing also the lights. In addition, we propose to change the markers to avoid repetitive appearances across consecutive frames.

\subsection{Future Purposes}
This paper is limited to one dataset and focuses on EfficientPose, but our future purposes include working on different existing \acrshort{6d} methods, in order to compare networks not only for their final ADD(-S) metric, but also for their generalization skills and their applicability.
Moreover, we plan to release the ArUco-Free Dataset, such that everyone could use it and measure generalization skills of their own model.

\section{Conclusions} \label{sec:5}
We analyzed the effect of the presence of ArUco markers in the dataset regarding the
learning procedure. In particular we found that they induce bias in the performance of the methodology. 
We assessed the generalization capabilities of one of the state-of-the-art \acrshort{6d} pose estimation networks, with respect to the two different datasets: with and without ArUco markers. The outcome was that the model trained on the dataset without markers achieves better results. However, we substituted ArUco markers with a black zone, and the black squares instead of markers are not enough to generalize.
To achieve a better generalization, the dataset should be augmented, with different lighting conditions, different views and more varied backgrounds. 
Our ArUco-Free LineMod dataset could be useful to prove more generalization capabilities of any newly proposed method.
In addition, the object's position, always in the center of the board is not beneficial from the generalization point of view, and it's preferable to crop the object once it's been detected.

Furthermore, the saliency analysis showed that the network definitely utilizes \acrshort{6d} pose information from visible markers and background objects.

This paper is the starting point to investigate the potential presence of bias in \acrshort{dnn} trained for \acrshort{6d} pose estimation, and to propose new methods to solve this problem.


\printglossary[type=\acronymtype, title=Abbreviations]

\section*{Availability of data and material}
\href{https://github.com/EleGo9/6DP-Data-Bias}{GitHub Repository}

\section*{Competing interests}
The authors declare that they have no competing interests.

\section*{Funding}
This research received no external funding.

\section*{Author's contributions}
All authors, EG and DS and CS and TP and GF and PA and MV and MB, contributed equally to the study conception and design. All authors, EG and DS and CS and TP and GF and PA and MV and MB, have read and agreed to the published version of the manuscript.

\section*{Acknowledgements}
This work was also supported by the  ``Gruppo Nazionale per il Calcolo Scientifico (GNCS-INdAM)''.

The publication was created with the co-financing of the European Union-FSE-REACT-EU, PON Research and Innovation 2014-2020 DM1062/2021.


\bibliographystyle{bmc-mathphys} 
\bibliography{bmc_article}      



\listoffigures








\end{document}